\documentclass[letterpaper, 10 pt, conference]{ieeeconf}
\IEEEoverridecommandlockouts

\usepackage[colorlinks]{hyperref}
\hypersetup{
    colorlinks,
    linkcolor=black,
    citecolor=black,
    filecolor=magenta,
    urlcolor=cyan,
}

\usepackage{resizegather}

\usepackage{cite}
\usepackage{amsmath,amssymb,amsfonts,mathrsfs}
\usepackage{graphicx}
\usepackage{textcomp}
\usepackage{xcolor}
\usepackage{tcolorbox}
\usepackage{textcomp}
\usepackage{tablefootnote}
\usepackage{threeparttable}
\usepackage{caption, subcaption}
\usepackage{bbm}
\usepackage{dsfont}
\usepackage{tikz}
\usepackage{graphicx}
\usepackage{tkz-euclide}
\usepackage{algorithm}
\usepackage{algpseudocode}
\usetikzlibrary{positioning}
\usetikzlibrary{shapes}
\usetikzlibrary{shapes.misc}
\usetikzlibrary{shapes.geometric}
\usetikzlibrary{plotmarks}
\usetikzlibrary{intersections}
\usetikzlibrary{calc}
\usetikzlibrary{fit}
\usetikzlibrary{patterns,tikzmark}
\usetikzlibrary{matrix,decorations.pathreplacing,calc}

\tikzset{cross/.style={cross out, draw, 
         minimum size=2*(#1-\pgflinewidth), 
         inner sep=0pt, outer sep=0pt}}

\newcommand{\vertiii}[1]{{\left\vert\kern-0.25ex\left\vert\kern-0.25ex\left\vert #1 
    \right\vert\kern-0.25ex\right\vert\kern-0.25ex\right\vert}}

\makeatletter
\renewcommand{\fps@figure}{htp}
\renewcommand{\fps@table}{htp}
\makeatother

\def\BibTeX{{\rm B\kern-.05em{\sc i\kern-.025em b}\kern-.08em
    T\kern-.1667em\lower.7ex\hbox{E}\kern-.125emX}}
    
\begin{document}

\title{Safety Certification in the Latent space using Control Barrier Functions and World Models}

\author{Mehul Anand$^{1}$, and
  Shishir Kolathaya$^{2}$ 
\thanks{$^{1}$ Indian Institute of Technology, Roorkee.
{\tt\scriptsize  mehul\textunderscore a@mt.iitr.ac.in}
.}
\thanks{$^{2}$ Center for Cyber-Physical Systems, Indian Institute of Science (IISc), Bengaluru.
{\tt\scriptsize  shishirk@iisc.ac.in}
.
}%
}

\newcommand{\mcom}[1]{{\color{red}{{[MT: #1]}}}}

\maketitle
\begin{abstract}
Synthesising safe controllers from visual data typically requires extensive supervised labelling of safety-critical data, which is often impractical in real-world settings. Recent advances in world models enable reliable prediction in latent spaces, opening new avenues for scalable and data-efficient safe control. In this work, we introduce a semi-supervised framework that leverages control barrier certificates (CBCs) learned in the latent space of a world model to synthesise safe visuomotor policies. Our approach jointly learns a neural barrier function and a safe controller using limited labelled data, while exploiting the predictive power of modern vision transformers for latent dynamics modelling.
\end{abstract}

\section{Introduction}
\label{section: Introduction}
As autonomous systems become increasingly prevalent, ensuring their safety remains a critical challenge—particularly when using learning-based controllers, which lack inherent guarantees. Various paradigms have been developed to address safety in control systems. Constrained Reinforcement Learning (CRL)~\cite{achiam2017constrained, NEURIPS2022_9a8eb202, NIPS2017_766ebcd5} encodes safety specifications as constraints during policy optimization, allowing data-driven learning but often lacking formal guarantees and risking unsafe behavior during exploration.

In contrast, optimal control approaches such as Hamilton-Jacobi (HJ) reachability~\cite{LYGEROS2004917, 8263977, tayal2025physics} provide rigorous safety analysis by characterizing safe sets through the solution of partial differential equations. However, their computational complexity scales poorly with state-space dimensionality, limiting their applicability to high-dimensional systems.

Control Barrier Functions (CBFs) have emerged as a scalable and effective tool for safety-critical control~\cite{Ames_2017, ames2019control, PRAJNA2006117, jagtap2020formal}. CBF-based methods often formulate the control synthesis as a Quadratic Program (QP), enabling real-time safety filtering~\cite{Ames_2017, ames2019control, pmlr-v120-taylor20a, tayal2023control}. However, traditional QP-based formulations are typically limited to control-affine systems and do not naturally handle input constraints. Certificate-based formulations~\cite{PRAJNA2006117, jagtap2020formal} extend the utility of CBFs by accommodating general nonlinear dynamics and bounded control inputs, making Control Barrier Certificates (CBCs) a versatile and scalable alternative.

A common limitation of these approaches is the reliance on known or approximate system dynamics. While approximate models may be available in state-feedback settings~\cite{9303847}, applying CBFs directly to visual observations remains nontrivial due to the absence of predictive models in the visual domain. Furthermore, synthesizing valid CBFs for arbitrary systems is itself a challenging task. Recent advances in neural CBFs~\cite{qin2022sablas, tayal2024learning} leverage the expressive power of neural networks to learn barrier certificates directly from data, enabling applicability to a broader range of systems and inputs, including vision.

Several recent efforts have explored the use of CBFs for safe control from visual observations~\cite{tong2023enforcingsafetyvisionbasedcontrollers, 10160805, 10610647}, but most of them rely on the assumption of control-affine dynamics, which limits generalization. Methods integrating Neural Radiance Fields (NeRFs) with CBFs~\cite{tong2023enforcingsafetyvisionbasedcontrollers} show promise for visuomotor control but incur significant computational overhead, impeding real-time deployment. Others use Generative Adversarial Networks (GANs)\cite{10160805} to infer 3D obstacle geometry for geometric CBF computation. Latent-space-based approaches\cite{harms2024neuralcontrolbarrierfunctions, tayal2024semi} generate CBFs from encoded visuomotor representations. However, these methods often employ autoencoders or GANs, which lack the capacity to model action-conditioned temporal dynamics, making them unsuitable for control and planning.

World models~\cite{zhou2025dinowmworldmodelspretrained} address this gap by learning structured, action-aware latent representations that capture system dynamics, enabling predictive planning and safe decision-making. In this work, we exploit the strengths of world models to synthesize safe visuomotor policies using learned control barrier certificates in the latent space.

\nocite{nakamura2025generalizingsafetycollisionavoidancelatentspace, tayal2025cpncbfconformalpredictionbasedapproach}

The main contributions of this work are:
\begin{itemize}
\item We propose a semi-supervised framework that integrates Control Barrier Certificates (CBCs) with a latent world model to enable safe visuomotor control using limited labeled data.
\item We demonstrate the effectiveness of our framework on two diverse systems—the Inverted Pendulum and Dubins Car—showing that safe policies can be synthesized without expert demonstrations.
\end{itemize}

Section~\ref{section: Background} reviews relevant background and problem formulation. Section~\ref{section: Methodology} outlines the proposed framework, detailing the world model architecture, neural barrier synthesis, and safe controller design. Section~\ref{section: Simulation} presents simulation results, and Section~\ref{section: Conclusions} concludes the paper with a discussion of limitations and future directions.

\section{Preliminaries}
\label{section: Background}
In this section, we will formally introduce Control Barrier Certificates (CBCs), and the problem formulation for the paper.

\subsection{System Description}

A discrete-time control system is a tuple $S = (X, U, f)$, where $X \subseteq \mathbb{R}^n$ is the state set of the system, $U \subseteq \mathbb{R}^{n_u}$ is the input set of the system, and $f : X \times U \to X$ describes the state evolution of the system via the following difference equation:
\begin{equation}
x(t + 1) = f(x(t), u(t)), \quad \forall t \in \mathbb{N},
\end{equation}
where $x(t) \in X$ and $u(t) \in U$,  denote the state and input of the system, respectively.

Consider a set S defined as the sub-zero level set of a continuous function $B : X \subseteq \mathbb{R}^n \to \mathbb{R}$ yielding,
\begin{align}
S &= \{x \in X \subset \mathbb{R}^n : B(x) \leq 0\} \\
X - S &= \{x \in X \subset \mathbb{R}^n : B(x) > 0\}.
\end{align}

We further restrict the class of S where its interior and boundary are precisely the sets given by $\text{Int}(S) = \{x \in X \subset \mathbb{R}^n : B(x) < 0\}$ and $\partial S = \{x \in X \subset \mathbb{R}^n : B(x) = 0\}$, respectively.

\subsection{Control Barrier Certificates (CBCs)}

In this section, we introduce the notion of a control barrier certificate, which provides sufficient conditions together with controllers for the satisfaction of safety constraints.

\textbf{Definition 1:} A function $B : X \to \mathbb{R}_0^+$ is a control barrier certificate for a discrete-time control system $S = (X, U, f)$ if for any state, $x \in X$ there exists an input $u \in U$, such that
\begin{equation}
B(f(x, u)) \leq B(x),
\end{equation}

The following lemma allows us to synthesise controllers for the discrete-time control system S, ensuring the satisfaction of safety properties.

\textbf{Lemma 1} (\cite{ames2019control}): For a discrete-time control system $S = (X, U, f)$, safe set $X_s \subseteq S$, and unsafe set $X_u \subseteq X - S$, the existence of a control barrier certificate, B, as defined in Definition 1, under a control policy $\pi : X \to U$ implies that the sequence state in S starting from $x_s \in X_s$ under the policy $\pi$ do not reach any unsafe states in $X_u$.

The zero-level set of the CBC $B(x) = 0$ separates the unsafe regions from the safe ones. For an initial state $x_0$ within the safe region ($x_0 \in X_s$), $B(x_0) \leq 0$ by condition (2). According to equation (4), which ensures $B(x)$ it remains non-increasing, the level set is not crossed, preventing access to unsafe regions. Therefore, ensuring system safety requires computing appropriate control barrier certificates and corresponding control policies.
\subsection{Problem Formulation}

Given a discrete-time robotic system S as defined in (1). Let $\mathcal{S}$ represent samples of visuomotor observations from the safe region, $\mathcal{U}$ represent samples from the unsafe region, and $\mathcal{D}$ denote the complete set of samples (both labelled and unlabeled). 

The objective is to devise an algorithm to jointly synthesise a provably correct parameterised barrier certificate $B_\theta$ and a safe parameterised policy $\pi_\theta$, with parameter $\theta$ such that it satisfies the condition (4) over the entire latent space defined by a world model, using a finite number of samples.

In the following section, we propose an algorithmic approach to solve the above problem.

\section{Methodology}
\label{section: Methodology}
We introduce a semi-supervised policy learning framework that employs control barrier certificates (CBCs) to jointly learn both the barrier certificate and a safe policy. This framework leverages the forward invariance properties of CBCs to ensure the learned policy remains within the safe set when initial conditions are safe. Since no barrier certificate exists initially, we start with initial safe and unsafe sets $\mathcal{X}_s \subseteq \mathcal{S}$ and $\mathcal{X}_u \subseteq \mathcal{X} - \mathcal{S}$ such that trajectories starting in $\mathcal{X}_s$ never enter $\mathcal{X}_u$, making the framework semi-supervised.

We collect datasets $\mathcal{C}$, $\mathcal{U}$, and $\mathcal{D}$ corresponding to $N$ visuomotor data points sampled from $\mathcal{X}_s$, $\mathcal{X}_u$, and the state set $\mathcal{X}$, respectively. The barrier function $B_\theta$ and policy $\pi_\theta$ are represented as neural networks learned in a latent space $z$ by a world model.

\subsection{Latent World Models}

The surge of excitement in large-scale foundation models has spurred the development of extensive video generation world models that are conditioned on agent actions, particularly in areas such as self-driving, control, and general-purpose video generation. These models are designed to produce video predictions based on textual descriptions or high-level action sequences. Although they have proven valuable for downstream applications such as data augmentation, their dependence on language-based conditioning restricts their effectiveness in scenarios requiring the achievement of visually specific goals. Furthermore, the reliance on diffusion models for video generation introduces significant computational overhead, limiting their suitability for test-time optimisation. Therefore, it makes sense to model the environment in a latent space rather than pixel space, as it will reduce computational requirements while retaining information about the environment.

Recent advances in world models for reinforcement learning, such as Dreamer v3~\cite{hafner2023mastering}, have demonstrated remarkable generality and performance across a wide range of tasks by learning environment dynamics in latent space and enabling agents to plan through imagined trajectories. Dreamer v3 employs a Recurrent State-Space Model (RSSM) that combines recurrent neural networks with stochastic latent variables to capture temporal dependencies and uncertainty in the environment. While this approach has led to impressive results, several shortcomings remain. Notably, the reliance on pixel-space reconstruction losses during training introduces significant computational overhead, especially when dealing with high-dimensional visual observations. This can hinder scalability and efficiency, particularly for long rollouts or large batch sizes. Additionally, the latent representations learned by RSSM, while expressive, may not always align with visually or semantically meaningful features required for visually grounded tasks, potentially limiting their effectiveness in scenarios demanding fine-grained visual understanding.

\begin{figure}
    \centering
    \includegraphics[width=1\linewidth]{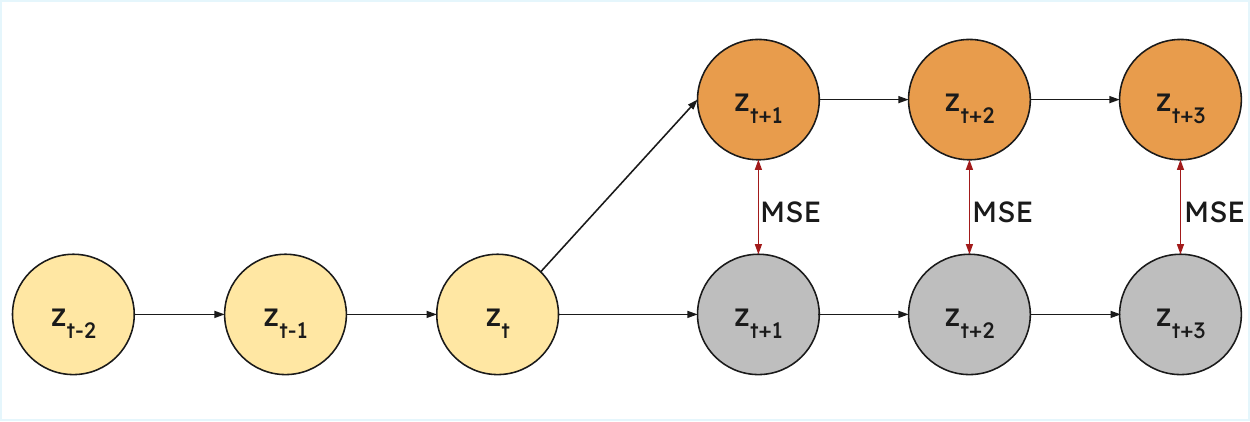}
 \caption{The Transition model is trained via autoregressive training on a latent consistency loss with a context length of 3 and a prediction horizon of 3; the actions provided are uniformly generated from an action space for generalisability of the dynamics}
    \label{fig:enter-label}
\end{figure}

The world model we are using consists of two parts, an observation model and a vision transformer to model the dynamics in latent space. The zero-shot encoder being used is a DINO-v2-small model\cite{oquab2024dinov2learningrobustvisual}, which gives us patch embeddings (16x16) and an embedded dimension of 384. Because DINO-v2 is a pre-trained model on a self-distillation loss, it allows the model to learn representations effectively, capturing semantic layouts and improving spatial understanding within images. The embeddings would have better separation of features than a custom-trained encoder. Hence, we learn the latent dynamics $d_\theta$ on a transition model, a transformer with a causal attention mask that takes a context length and proprioceptive information from the actual environment.

\begin{figure*}[t!]
    \centering
    \includegraphics[width=0.85\textwidth]{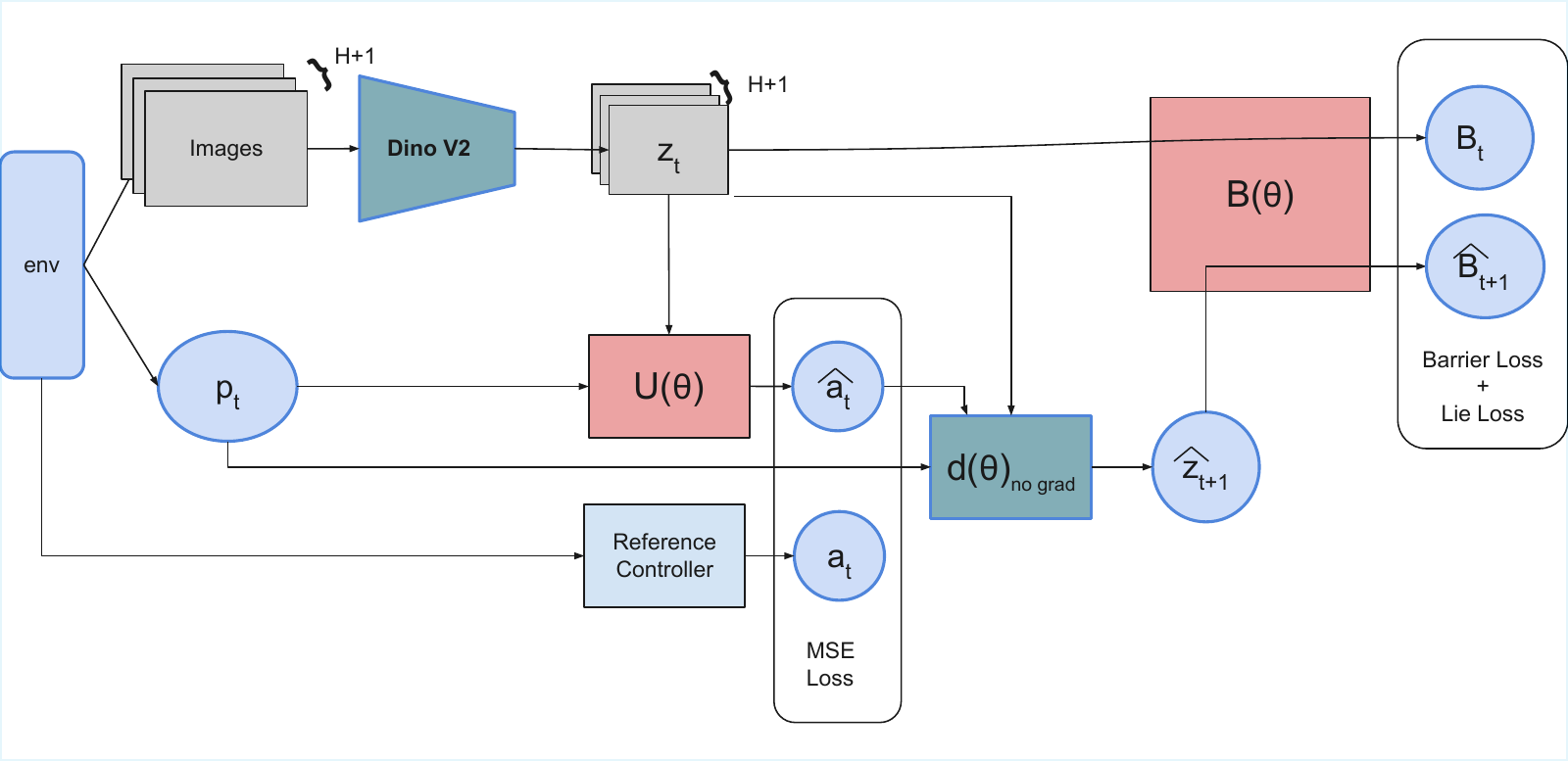}
    \caption{The figure encapsulates the flow of our entire framework, the environment  with a reference controller generates the image at time t, and proprioceptive information. The image is then encoded by Dino v2, which is further stacked onto a context of length H. This is used to predict the next state by the action given by our controller U, which is then further fed into the barrier function to check if it satisfies the lie condition}
    \label{fig:enter-label}
\end{figure*}

More specifically, at each time step t, our world model consists of the following components:
\begin{align}
\text{Observation model: } z_t \sim \mathbf{enc} (z_t | o_t )
\end{align}
\begin{align}
\text{Transition model: } z_{t+1} \sim \mathbf{d_\theta}(z_{t+1} | z_{t-H:t}, a_{t-H:t}, p_{t-H:t} )
\end{align}
where the observation model encodes image observations to latent states $z_t$, and the transition model takes in a history of past latent states of length H. $a_t$ is the action taken at time step t and $p_t$ is the proprioceptive information at time step t.\cite{zhou2025dinowmworldmodelspretrained}

\begin{equation}
L_{\text{pred}} = \|d_\theta (\text{enc}_\theta (o_{t-H:t} ), a_{t-H:t} , p_{t-H:t} ) -\text{enc}_\theta (o_{t+1} )\|
\end{equation}

\subsection{Barrier Network}

Since we are working with a latent space, the only way to formulate a barrier function is via a neural network with $\tanh$ activation functions and an unbounded final layer. Because of how well DINO-v2 captures spatial information, the states are already well separated in the latent dimension. This makes it easier to learn the difference between safe and unsafe states; hence, it can be done with a small network. The separability of safe/unsafe regions in feature space is crucial for barrier certificate accuracy. We use a loss that learns a latent representation where safe/unsafe regions are distinctly separable.

The loss function integrates safety segregation:
\begin{align}
L_{\text{barrier}}(\theta) &= \xi_1 \sum_{z_i \in \mathcal{C}} \max(0, B_\theta(z_i) ) \nonumber \\
&+ \xi_2 \sum_{z_i \in \mathcal{U}} \max(0, -B_\theta(z_i) ) \nonumber \\
\end{align}
where $\xi_1, \xi_2 > 0$ are the weighting coefficients. This design ensures that safe states are mapped to non-positive barrier values, while unsafe states are mapped to non-negative values, thereby enforcing a clear separation in the latent space.

To further regularise the barrier and encourage smooth transitions between states, we introduce the Lie loss:
\begin{align}
L_{\text{lie}}(\theta) &=  \sum_{z_{i+1} \in \mathcal{C}} \max(0, B_\theta(z_{i+1})- \alpha B_\theta(z_{i})) \nonumber \\
&+  \sum_{z_{i+1} \in \mathcal{U}} \max(0, B_\theta(z_{i})- \alpha B_\theta(z_{i+1}) ) \nonumber \\
\label{eq:lie}
\end{align}
where $\alpha >0 $

The Lie loss not only smoothens the barrier function but also encodes the principle that lower barrier values correspond to safer regions. This encourages monotonicity in the barrier function along safe trajectories and penalises decreases along unsafe ones, thereby improving the robustness of safety guarantees.

The barrier network takes the current state $z_t$ as input and outputs a scalar value representing the safety certificate. While it is possible to augment the input with proprioceptive information, this would require the transition function to predict proprioceptive values for the next state, which is unnecessary for our simplified environments.

\subsection{Controller Synthesis}
\label{subsec:controller}

Similar to the barrier network, the controller network is also a feed-forward neural network, but its outputs are bounded by a sigmoid function to ensure valid action ranges. The controller receives as input the current latent state $z_t$ and current proprioceptive information $p_t$, and outputs the action $a_t$. All hidden layers employ $\tanh$ activation functions to preserve the ability to represent both positive and negative values.

Based on Lemma 1, satisfying condition (4) ensures trajectories starting in $\mathcal{X}_s$ avoid $\mathcal{X}_u$. The synthesis loss enforces this condition even in the presence of latent dynamics errors:
\begin{equation}
L_{\text{syn}}(\theta) = \sum_{x_i \in \mathcal{D}} \max\left(0, B_{\theta}(d_\theta(z_t, \pi_\theta(z_t), p_t)) - B_\theta(z_t) \right)
\end{equation}
This loss, similar in spirit to the Lie loss in equation (\ref{eq:lie}), encourages the controller to select actions that do not increase the barrier function, thereby maintaining safety over time.

To further enhance performance while maintaining safety, we introduce an MSE loss to a user-defined policy $\pi_{\text{user}}$:
\begin{equation}
L_\pi(\theta) = \|\pi_\theta - \pi_{\text{user}}\|^2
\end{equation}
This term ensures that the learned controller remains close to a desired behaviour, which can encode preferences or prior knowledge. Importantly, our approach does not require expert demonstrations. This flexibility allows for scalable and data-efficient synthesis of safe controllers in complex, high-dimensional environments.

\subsection{Training Scheme}
\label{subsec:training}
The training procedure is structured into two distinct stages:

\begin{enumerate}
    \item \textbf{World Model Training}: We begin by training the world model using a dataset of 50,000 state transitions generated from random actions across both environments. The model is trained with a context window of length 3 and a prediction horizon of 3, enabling it to capture short-term temporal dependencies and accurately forecast future states. 

    \item \textbf{Barrier and Controller Training}: After establishing the world model, we collect approximately 250 trajectories with random initialisation using a reference policy to provide diverse coverage of the state space. Both the barrier network and the controller are trained simultaneously during this stage. To promote generalisation and prevent overfitting, training of the barrier network is halted once its loss converges. The controller, however, continues to be optimised until its own convergence criterion is met. This staged approach ensures that the barrier remains a robust safety certificate while allowing the controller to fully exploit the learned safe regions for effective and safe policy execution.
\end{enumerate}

\begin{algorithm}
\caption{Safe Controller Synthesis}
\label{alg:training}
\begin{algorithmic}[1]
\Require{Datasets $\mathcal{C}, \mathcal{U}, \mathcal{D}, \mathbf{d_\theta}, enc, B_\theta, \pi_\theta, \pi_\text{user}$}
\State Initialise $\theta$
\For{$i = 1$ \textbf{to} \text{maxiter}}
    \State $\mathcal{L_\text{pred}}  \xleftarrow{}(d_\theta, enc, O_{t-H:t+1}, a_{t-H:t},p_{t-H:t})$
    \State $\theta \xleftarrow{} \text{Learn }  \theta$ 
    \State $d_\theta \xleftarrow{} \theta$
\EndFor
\State Initialise $\theta$
\While{until Convergence}
    \State $\mathcal{L_\text{barrier}}  \xleftarrow{}(B_\theta, \mathcal{C}, \mathcal{U}, \mathcal{D})$
    \State $\mathcal{L_\text{lie}}  \xleftarrow{}(B_\theta, \mathcal{C}, \mathcal{U})$
    \State $\mathcal{L_\text{syn}}  \xleftarrow{}(B_\theta, \pi_\theta,d_\theta, \mathcal{D})$
    \State $\mathcal{L_\pi}  \xleftarrow{}( \pi_\theta, \pi_\text{user})$
    \State $\mathcal{L_\text{total}} = \mathcal{L_\text{barrier}} + \mathcal{L_\text{lie}}+\mathcal{L_\text{syn}} +\mathcal{L_\pi}$
    \State  $\theta \xleftarrow{} \text{Learn }  \theta$ 
    \State $B_\theta, \pi_\theta \xleftarrow{} \theta$
\EndWhile
\end{algorithmic}
\end{algorithm}

\section{Simulations}
\label{section: Simulation}
In this section, we assess the efficacy of our proposed framework through two distinct case studies: the inverted pendulum system and the obstacle avoidance of a Dubin's car. Both case studies are conducted on a computing platform equipped with an Intel i9-11900K CPU, 32GB RAM, and NVIDIA GeForce RTX 4090 GPU.

\section*{A. Inverted Pendulum}
We analyse an inverted pendulum system characterised by the state vector $\mathbf{x} = [\Theta, \dot{\Theta}] \in [-\pi, \pi] \times [-3.5, 3.5]$, where $\Theta$ represents the angular position, and $\dot{\Theta}$ denotes the angular velocity. The control input to the system is the applied torque $\tau \in [-6, 6]   \text{ Nm}$, as illustrated in Fig. 2a. The discrete-time dynamics governing this system are given by:

\begin{equation}
\begin{bmatrix}
\Theta_{t+1} \\
\dot{\Theta}_{t+1}
\end{bmatrix}
=
\begin{bmatrix}
\Theta_t \\
\dot{\Theta}_t
\end{bmatrix}
+
\left(
\begin{bmatrix}
\dot{\Theta}_t \\
\frac{g}{l} \sin(\Theta_t)
\end{bmatrix}
+
\begin{bmatrix}
0 \\
\frac{1}{ml^2}
\end{bmatrix} u_t
\right) \Delta t
\label{eq:14}
\end{equation}

\begin{figure}[htbp]
    \centering
    \begin{subfigure}{0.4\linewidth}
        \centering
        {\includegraphics[width=\linewidth]{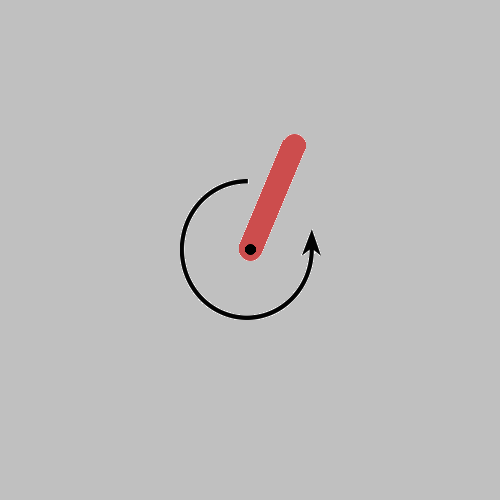}}
        \caption{}
        \label{fig:sub1}
    \end{subfigure}
    \hfill
    \begin{subfigure}{0.5\linewidth}
        \centering
        {\includegraphics[width=\linewidth]{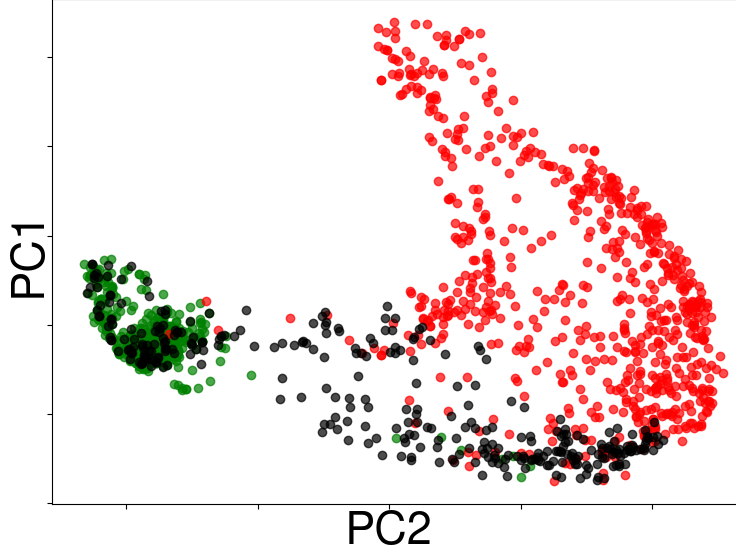}}
        \caption{}
        \label{fig:sub2}
    \end{subfigure}
    \caption{\textbf{a}. represents the image of the inverted pendulum in the OpenAI gymnasium environment. \textbf{Fig b} The embedding, when projected onto the first two principal components, enables a clear visualisation of the decision boundary that partitions safe and unsafe regions. This separation is achieved with a low rate of misclassified points. }
    \label{fig:main}
\end{figure}
where $m$ and $l$ represent the mass and length of the pendulum, respectively.

The datasets, $\mathcal{C}$, $\mathcal{U}$, and $\mathcal{D}$, are sampled as follows:

\begin{equation}
\begin{aligned}
\mathcal{C} &= \left\{ O_t \mid \mathbf{x}_t \in \left[ -\frac{\pi}{12}, \frac{\pi}{12} \right] \times [-0.25, 0.25] \right\} \\
\mathcal{U} &= \left\{ O_t \mid \mathbf{x}_t \in [-\pi, \pi] \times [-3.5, 3.5] \setminus \left[ -\frac{\pi}{2}, \frac{\pi}{2} \right] \times [-1.5, 1.5] \right\} \\
\mathcal{D} &= \left\{ O_t \mid \mathbf{x}_t \in [-\pi, \pi] \times [-3.5, 3.5] \right\}
\end{aligned}
\label{eq:16}
\end{equation}

Here $O_t$ is a stack of the latent state representation of the RGB frames of the environment and $\dot{\Theta}$ at time t.

A reference policy, a simple PD controller, is used to roll out data for imitation $\pi_{\text{ref}}$.

Visualisations of the trained barrier function are presented in both the latent space with sample points (Fig. 2b) and in the plane (Fig. 2c). These visualisations demonstrate the successful separation of the safe region from the unsafe region. As shown in Fig. 2d, the trajectories generated by the learned policy $\pi_\theta$ start in the unsafe region and successfully reach the safe region, validating our approach  

\section*{B. Obstacle avoidance using Dubins' Car}

We set up an environment with an obstacle of fixed radius at the centre and an end goal that the agent has to reach. The agent needs to learn a controller to reach the end goal while avoiding the obstacle. The reference controller is a simple P controller.   

The state vector is defined as $[x_t , y_t , \Theta_t ]^T\in [-1.5, 1.5]^2 \times [-\pi, \pi]$, where $(x_t , y_t)$ denotes the robot’s position and $\Theta_t$ represents its orientation.  The agent moves at a constant speed $v$=1, while the control input $u$ corresponds to $\dot{\Theta}$ turning the agent about its axis.
\begin{figure}[htbp]
    \centering
    \begin{subfigure}{0.4\linewidth}
        \centering
        {\includegraphics[width=\linewidth]{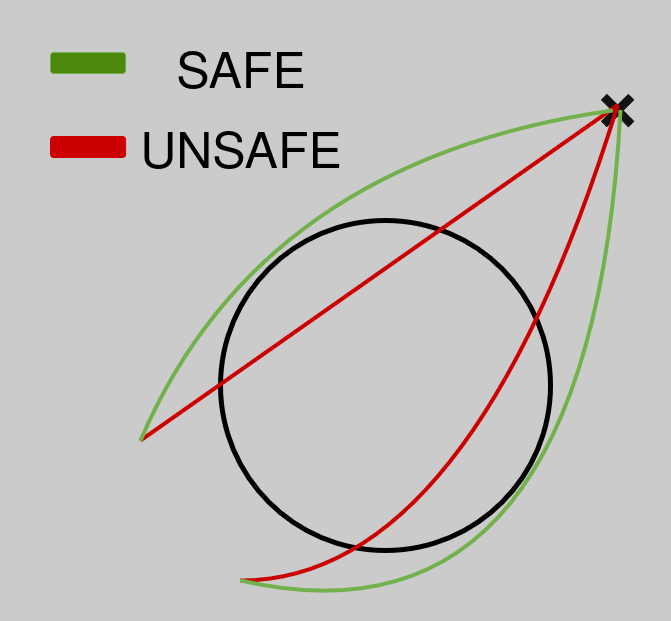}}
        \caption{}
        \label{fig:sub1}
    \end{subfigure}
    \hfill
    \begin{subfigure}{0.5\linewidth}
        \centering
        {\includegraphics[width=\linewidth]{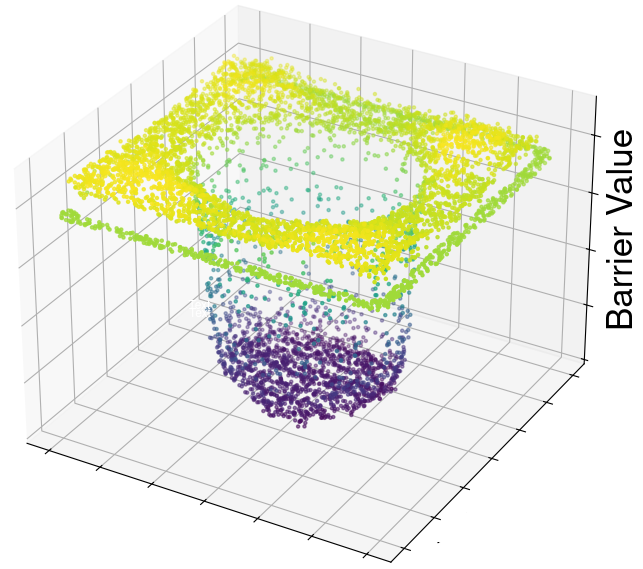}}
        \caption{}
        \label{fig:sub2}
    \end{subfigure}
    \caption{\textbf{a} This figure shows the difference between the trajectories generated by our controller (green) and the reference controller(red). \textbf{Fig b} represents the value of barrier function over the state space of the env}
    \label{fig:main}
\end{figure}
It follows the discrete-time dynamics given by:
\begin{align}
\begin{bmatrix}
\mathbf{x}_{t+1} \\
\mathbf{y}_{t+1} \\
\Theta_{t+1}
\end{bmatrix}
&=
\begin{bmatrix}
\mathbf{x}_t \\
\mathbf{y}_t \\
\Theta_t
\end{bmatrix}
+
\left(
\begin{bmatrix}
v \cos \Theta_t \\
v \sin \Theta_t \\
0
\end{bmatrix}
+
\begin{bmatrix}
0 \\
0 \\
1
\end{bmatrix}
\mathbf{u}_t
\right) dt
\end{align}
The datasets, $\mathcal{C}$, $\mathcal{U}$, and $\mathcal{D}$, are sampled as
\begin{equation}
\begin{aligned}
\mathcal{C} &= \left\{ O_t \mid \mathbf{x}_t \in [ -1.5, 1.5]^2 \times [-\pi, \pi] \setminus [-0.9, 0.9]^2 \times [-\pi, \pi] \right\} \\
\mathcal{U} &= \left\{ O_t \mid \mathbf{x}_t \in [-0.7, 0.7]^2 \times [-\pi, \pi] \right\} \\
\mathcal{D} &= \left\{ O_t \mid \mathbf{x}_t \in [-1.5, 1.5]^2 \times [-\pi, \pi] \right\}
\end{aligned}
\label{eq:16}
\end{equation}
     Here $O_t$ is a stack of the latent state representation of the RGB frames of the environment and the heading angle at time t.

\section{Conclusions}
\label{section: Conclusions}
In this work, we presented a semi-supervised framework for synthesising safe visuomotor policies by leveraging control barrier certificates (CBCs) in the latent space of a world model. Our approach enables the joint learning of both a neural barrier function and a safe controller using limited labelled data, while exploiting the predictive capabilities of modern vision transformers for latent dynamics modelling. Through case studies on the inverted pendulum and Dubins' car environments, we demonstrated that our method can effectively separate safe and unsafe regions in the latent space and synthesise policies that promote safe behaviour.

\textbf{Limitations and Future Work}: While our results are promising, the current framework does not provide formal safety guarantees; it instead encourages safer behaviour via the learned barrier. Future work will focus on extending this framework to provide formal verification of safety properties, scaling to higher-dimensional and more complex environments, and exploring integration with more expressive world models. We believe that combining neural CBCs with powerful world models paves the way for scalable, data-efficient, and provably safe visuomotor control in real-world robotics applications.

\label{section: References}
\bibliographystyle{IEEEtran}
\bibliography{references.bib}

\end{document}